# Challenge report: VIPriors Action Recognition Challenge

Zhipeng Luo, Dawei Xu, Zhiguang Zhang
DeepBlue Technology (Shanghai) Co., Ltd
{ luozp, xudw, zhangzhg}@deepblueai.com

*Abstract*—**This paper is a brief report to our submission to the VIPriors Action Recognition Challenge. Action recognition has attracted many researchers' attention for its full application, but it is still challenging. In this paper, we study previous methods and propose our method. In our method, we are primarily making improvements on the SlowFast Network and fusing with TSM to make further breakthroughs. Also, we use a fast but effective way to extract motion features from videos by using residual frames as input. Better motion features can be extracted using residual frames with SlowFast, and the residual-frame-input path is an excellent supplement for existing RGB-frame-input models. And better performance obtained by combining 3D convolution(SlowFast) with 2D convolution(TSM). The above experiments were all trained from scratch on UCF101.**

## I. Introduction

Video-based action recognition has lately received considerable attention for its practicability in many areas. Numerous work has been carried out in this area from the academic community[1], [3], [4]. In action recognition, there are currently three mainstream approaches – Two-Stream Convolutional Networks[5], CNN+LSTM and 3D convolution[6]. The two-stream networks mainly extract RGB information and dense optical flow between every two frames. It is troublesome to extract the optical flow characteristics for the large dataset in the competition. Then we use some improvements to replace the optical flow[7], [8], [9]. TSN[10] and the improved network inherit the structure of the two-stream network. There are some improvements between temporal-spatial fusion and the relation of temporal sequence. Considering the difficulty of temporal feature extraction, some networks specifically designed to deal with temporal features, such as LSTM[12], which is also effective for action recognition. The CNN+LSTM[11] networks have been employed for modeling temporal dynamics, which uses a CNN to extract frame features and an LSTM to integrate features over time. C3D[12], [13], [14] is a simple and efficient 3D convolution network. However, compared to 2D convolution[15], 3D convolution on sequences of dense frames are computationally expensive, and the model size also has a quadratic growth.

We are primarily making improvements on the SlowFast Network and fusing with TSM to make further breakthroughs. At the same time, we combined RGB and residual-frame as SlowFast inputs to enable the network to extract more dynamic characteristics.

## II. Network

Video contains a lot of time and space information. We need frame-level feature representation and temporal modeling to fuse these information. SlowFast can be used to extract dynamic information with 3D ConvNets. In addition, when treating each frame of a video as a single image, TSM can be used as a supplement to spatial information.

The SlowFast model includes a slow pathway, working at a low frame rate, to capture spatial semantics. And a Fast path, working at a high frame rate, to capture motion at fine temporal resolution. The Fast pathway can be made very lightweight by reducing its channel capacity, yet it can learn useful temporal information for video recognition.

In SlowFast, we tried different backbones including ResNet-50 and ResNet-101, optionally augmented with non-local (NL) blocks. We combined the network structure of 2+1D network, and one $k \times k \times k$ kernel can be separated into three parts, $k \times 1 \times 1$, $1 \times k \times k$ and $k * k * k$. Meanwhile, we tried the SE module and the TAM module. And attempt to add Spatio-temporal attention to 3D convolution.

In contrast to existing 3D convolution-based methods that use ReLU as the activation function, replacing ReLU with ELU improved the accuracy. And we replace all the downsampling convolution layers with max-pooling layers(POOL). For the input of SlowFast, we combined residual frames and RGB frames to form a two-path solution, which can get the best effect.

Because residual frames contain little information of object appearance, we further use a 2D convolution TSM to extract appearance features.

Temporal Shift Module (TSM) are variants of 2D ConvNets, which can be seen in Figure 3. TSM shifts part of the channels along with the temporal dimension, facilitating information exchange among neighboring frames. It can be inserted into 2D CNNs to achieve temporal modeling at zero computation and zero parameters. Shift operation is actually used to integrate the frame features coming from the front and rear frames to increase the time perception field. We choose the embedding method of residual TSM to ensure the extraction of spatiotemporal features without increasing the computation.

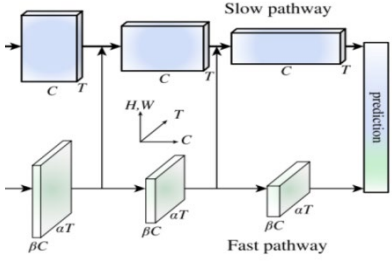

**Figure 1. SlowFast Network**

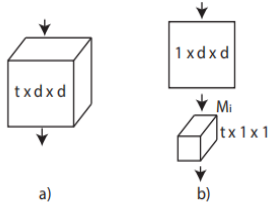

**Figure 2. (2+1)D convolutional block splits the computation into a spatial 2D convolution followed by a temporal 1D convolution**

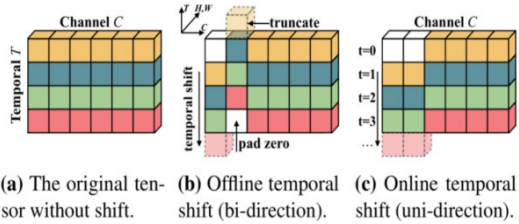

**Figure 3. Temporal Shift Module.**

## III. EXPERIMENT

### A. Data augmentation

In this competition, we take 64 consecutive frames as the input of SlowFast. Since training from scratch, we further enhance the data diversity. We do multi-scale enhancements by randomly scale the data to 0.8x-1.25x of 112, then randomly crop each frame to 112*112. In addition to lighting, horizontal flipping, corner cropping, contrasting in image enhancement, we play the video in reverse order as new training samples. And randomly extracting frames in one or double step. The method of data enhancement described above applies to the residual-frame-input path. The residual-frame-input path is a good supplement for existing RGB-frame-input models. The input of TSM is 16 consecutive frames, and the same method as above is used for data enhancement.

**Table 1: The results when use different modules.**

| Method | Acc(%) |
| --- | --- |
| SlowFast 3DResNet 50 | 70.0 |
| SlowFast +non_local | 70.6 |
| SlowFast +2plus1D | 71.2 |
| SlowFast +ELU+POOL | 73.1 |
| SlowFast +ELU+POOL+non_local | 72.8 |
| SlowFast +ELU+POOL+TAM | 71.2 |
| SlowFast +ELU+POOL+(3*3+3*1+1*3) | 73.5 |
| SlowFast +attention | 70.5 |

### B. Training

In the experiment, we tested MFnet, R2plus1D, ResNetR3D, SlowFast network, and TSM networks, and finally chose to use SlowFast network and TSM for model fusion. The structure of SlowFast and 2+1D can be seen in Figure 1 and Figure 2 respectively. In experiments, MFnet, R2plus1D, and ResNetR3D can only get low accuracy compared to SlowFast. It is very difficult to train TSM from scratch because of the extra shift module which can be shown in Figure 3. But TSM can make full use of single-frame information and the fusion between channels can perfectly cooperate with SlowFast multi-frame fusion information.

For it is difficult to train TSM. We first used single-frame as input to train TSM, when the accuracy of the single-frame training method model reaches 53%, we fine-tune TSM by using 16-frames as input.

We adopted the same train strategy as TSM train SlowFast. After the accuracy of SlowFast reaches 50%, we saved it as a pre-trained model. Based on this pre-trained model we do more research, such as add 3D convolution modules, non_local and 2plus1D modules. And adopted data augmentation to fun-tune it. Also fun-tune it by using normal and reverse contacted video. As shown in Table 1. Finally, we choose SlowFast with ELU, POOL, and 2+1D network with k=3 as our modified SlowFast.

In addition to adding modules to the model, we've tried to use 3DResNet50 or 3DResNet101 as the backbone of the model respectively. The result as shown in Table 2, and the performance of 3DResNet101 is not as good as that of 3DResNet50, so we choose 3DResRet50 as the backbone.

We use the standard SGD with momentum set to 0.9 in all cases. The batch size is set to 128. In the experiment, the learning rate was set to 0.001, and weight decay is set to 0.0005 at the beginning. The learning rate is reduced to 1/10 for every twenty epochs.

Finally, by using Residual frames or RGB frames respectively as input, we trained four models with the best SlowFast structure through the data augmentation mentioned above.

**Table 2: The results when use different backbone**

| Method | Input type | Acc(%) |
| --- | --- | --- |
| SlowFast+ELU+POOL+(3*3+3*1+1*3) 3D ResNet 50 | RGB | 73.5 |
| SlowFast+ELU+POOL+(3*3+3*1+1*3) 3D ResNet 101 | RGB | 72.1 |
| SlowFast+ELU+POOL+(3*3+3*1+1*3) 3D ResNet 50 | Diff | 72.8 |
| SlowFast+ELU+POOL+(3*3+3*1+1*3) 3D ResNet 101 | Diff | 71.9 |
| TSM(with shift) 2D ResNet 50 | RGB | 72.1 |
| TSM(with shift) 2D ResNet 101 | RGB | 70.8 |
| TSM(no_shift) 2D ResNet 50 | RGB | 71.9 |
| TSM(no_shift) 2D ResNet 101 | RGB | 71.2 |

## C. Test

After get trained model based on above method. In test stage, we adopted test-time augmentation to get better result. We conducted different test fusion methods based on the above models. In all tests, we randomly select 64 consecutive frames from the video as input, and take ten consecutive times to calculate the average as the final score of the entire video. Note: we use the data before Softmax for ensemble. As shown in Table 3.

**Table 3: The results when use different test augmentation.**

| Network | Input type | Test augmentation | Acc(%) | ID |
|---|---|---|---|---|
| SlowFast | RGB Interval 1 | Center crop | 75.5 | 1 |
| SlowFast | Diff Interval 1 | Center crop | 74.6 | 2 |
| SlowFast | RGB Interval 1 | Horizontal Flip | 75.6 | 3 |
| SlowFast | Diff Interval 1 | Horizontal Flip | 74.3 | 4 |
| SlowFast | RGB Interval 1 | Random crop | 73.3 | 5 |
| SlowFast | RGB Interval 1 | Normal and reverse contacted video | 75.1 | 6 |
| SlowFast | RGB Interval 1 | Reverse order | 74.6 | 7 |
| SlowFast | Diff Interval 1 | Normal and reverse contacted video | 72.1 | 8 |
| SlowFast | Diff Interval 1 | Reverse order | 71.6 | 9 |
| TSM | RGB Interval 1 | Center crop | 70.8 | 10 |
| TSM | RGB Interval 2 | Center crop | 71.2 | 11 |
| TSM | RGB Interval 1 | Random crop | 71.3 | 12 |
| TSM | RGB Interval 1 | Horizontal Flip | 72.2 | 13 |
| TSM | RGB Interval 1 | Reverse order | 70.8 | 14 |
| SlowFast | RGB Interval 2 | Center crop | 71.3 | 15 |
| SlowFast | Diff Interval 2 | Center crop | 74.5 | 16 |
| SlowFast | RGB Interval 2 | Horizontal Flip | 75.4 | 17 |
| SlowFast | Diff Interval 2 | Horizontal Flip | 74.0 | 18 |
| SlowFast | RGB Interval 2 | Random crop | 73.2 | 19 |
| SlowFast | RGB Interval 2 | Normal and reverse contacted video | 75.0 | 20 |
| SlowFast | RGB Interval 2 | Reverse order | 73.6 | 21 |
| SlowFast | Diff Interval 2 | Normal and reverse contacted video | 72.0 | 22 |

To get the final result, we ensemble results that got above in different methods. The ID n in Table 4 correspond to ID n in Table 3. Finally, we got our best score of 87.6%, as shown in Table 4.

## IV. CONCLUSION

In this paper, we mainly focused on extracting motion features without optical flow. 3D ConvNets are believed to be capable of capturing motion features when combining RGB and residual frames as the network input. The overhead for this computation was negligibly small. Residual frames can be a fast but effective way for a network to capture motion features, and they are a good choice for avoiding complex computation for optical flow.

Finally, We combined modified SlowFast and TSM, using both action information and space-time information for optimal results.

**Table 4: The result of different ensemble method**

| Ensemble method | Acc(%) |
|---|---|
| (SlowFast)(Interval 1) RGB+Diff Center crop + Horizontal Flip+ Random crop+ Concat normal and reverse order | 84.5 |
| (SlowFast +TSM)(Interval)RGB+Diff center crop+ Horizontal Flip+ Random crop+ Concat normal and reverse order + Reverse order | 86.3 |
| (SlowFast +TSM)(Interval +2)RGB+Diff Center crop *2+ Horizontal Flip*2+ Random crop+ Concat normal and reverse order + Reverse order | 87.6 |
| (SlowFast +TSM)(Interval+2)RGB+Diff center crop*4+ Horizontal Flip*4+ Random crop*2+ Concat normal and reverse order + Reverse order | 85.8 |